\documentclass[twoside]{article}

\usepackage[accepted]{aistats2020}

\usepackage[round]{natbib}

\usepackage[utf8]{inputenc} 
\usepackage[T1]{fontenc}    
\usepackage{url}            
\usepackage{booktabs}       
\usepackage{amsfonts}       
\usepackage{nicefrac}       
\usepackage{microtype}      

\usepackage{amsmath}
\usepackage{amsthm}
\usepackage{amssymb}
\usepackage{enumitem}

\usepackage[titletoc, toc]{appendix}

\usepackage{graphicx}
\graphicspath{ {./fig/} }

\usepackage[bf]{caption}
\usepackage{float}
\usepackage{xcolor}

\usepackage{color}
\definecolor{mydarkblue}{rgb}{0,0.08,0.45}
\definecolor{mygreen}{rgb}{0.3, 0.69, 0.29}
\definecolor{darkred}{rgb}{0.6, 0, 0.05}
\usepackage[colorlinks=true,allcolors=mydarkblue]{hyperref}

\newcommand{\R}{\mathbb{R}}

\newcommand{\N}{\mathcal{N}}

\newcommand{\Dir}{\text{Dirichlet}}
\newcommand{\Cat}{\text{Categorical}}

\newcommand{\boldY}{\mathbf{Y}}

\newcommand{\boldI}{\mathbf{I}}

\newcommand{\boldy}{\mathbf{y}}

\newcommand{\boldw}{\mathbf{w}}
\newcommand{\boldz}{\mathbf{z}}

\newcommand{\boldzero}{\mathbf{0}}

\newcommand{\boldpi}{\boldsymbol{\pi}}

\newcommand{\boldalpha}{\boldsymbol{\alpha}}

\newcommand{\calL}{\mathcal{L}}

\setlist[itemize]{noitemsep, topsep=0pt}

\begin{document}

\setlength{\abovedisplayskip}{4pt}
\setlength{\belowdisplayskip}{4pt}

\runningtitle{BasisVAE: Translation-invariant feature-level clustering with Variational Autoencoders}

\twocolumn[

\aistatstitle{BasisVAE: Translation-invariant feature-level clustering \\with Variational Autoencoders}

\aistatsauthor{Kaspar M\"artens \And  Christopher Yau} 

\aistatsaddress{University of Oxford \And Alan Turing Institute\\University of Birmingham\\University of Manchester} 
]

\begin{abstract}
Variational Autoencoders (VAEs) provide a flexible and scalable framework for non-linear dimensionality reduction. However, in application domains such as genomics where data sets are typically tabular and high-dimensional, a black-box approach to dimensionality reduction does not provide sufficient insights. Common data analysis workflows additionally use clustering techniques to identify groups of similar \emph{features}. This usually leads to a two-stage process, however, it would be desirable to construct a joint modelling framework for simultaneous dimensionality reduction and clustering of features. In this paper, we propose to achieve this through the \emph{BasisVAE}: a combination of the VAE and a probabilistic clustering prior, which lets us learn a one-hot basis function representation as part of the decoder network. Furthermore, for scenarios where not all features are \emph{aligned}, we develop an extension to handle translation-invariant basis functions. We show how a collapsed variational inference scheme leads to scalable and efficient inference for BasisVAE, demonstrated on various toy examples as well as on single-cell gene expression data. 
\end{abstract}

\section{Introduction}

\begin{figure*}[!ht]
    \centering
    \includegraphics[width=0.9\textwidth]{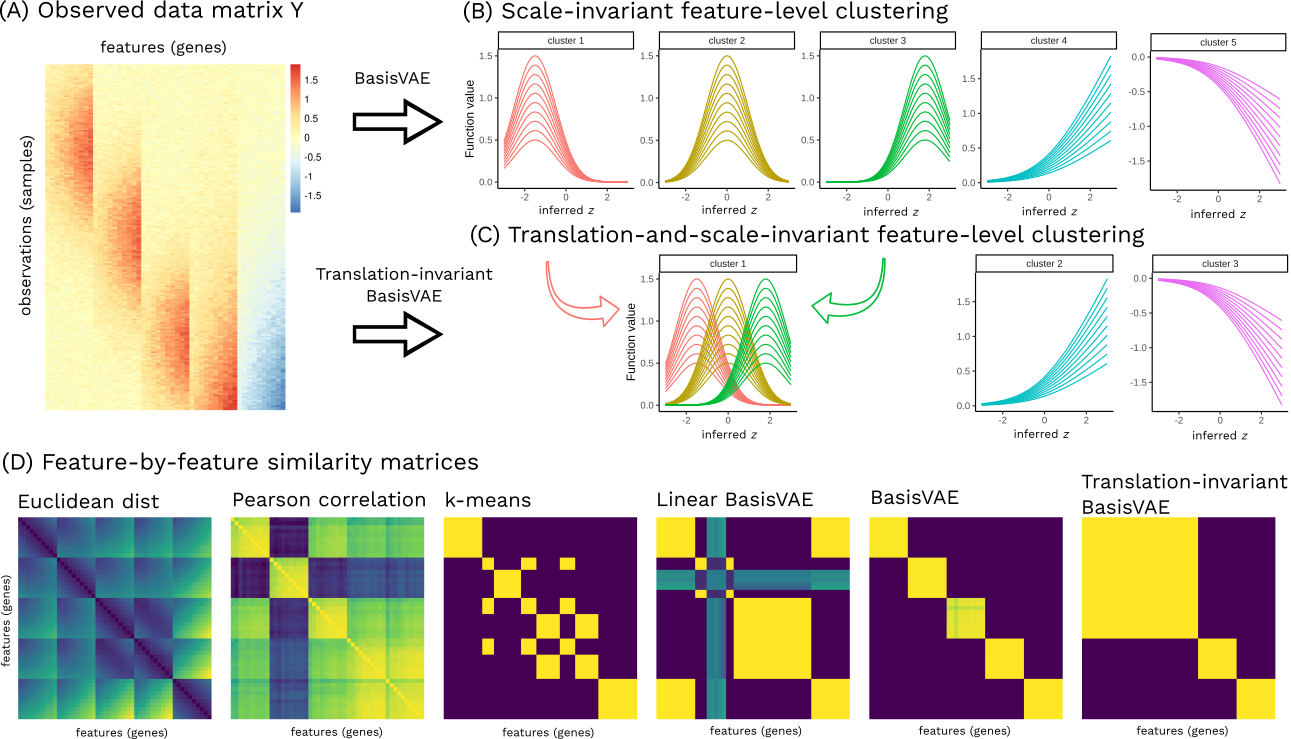}
    \caption{Illustration of our goal: Given a high-dimensional data matrix $\boldY$ (heatmap in panel (A)), we want to learn a low-dimensional representation $\boldz$ and simultaneously cluster features based on the mappings $\boldz \mapsto \boldY$ (to make it easier for the reader, we have ordered rows in panel (A) according to the inferred $\boldz$). Panels (B) and (C) show the inferred clusters: clustering (B) is scale-invariant whereas clustering (C) is additionally translation-invariant. For every feature we have plotted the inferred neural network mapping $\boldz \mapsto \boldy^{(j)}$. These have been grouped into subpanels which correspond to inferred clusters). (D) shows feature-by-feature similarity matrices either based on distances (Euclidean, Pearson correlation), or co-clustering metrics either for k-means (with $K=5$) or co-occurrence posterior probabilities in the BasisVAE framework.
    }
    \label{fig:illustration}
    \vspace{-3mm}
\end{figure*}

\begin{figure}[!ht]
    \centering
    \includegraphics[width=\columnwidth]{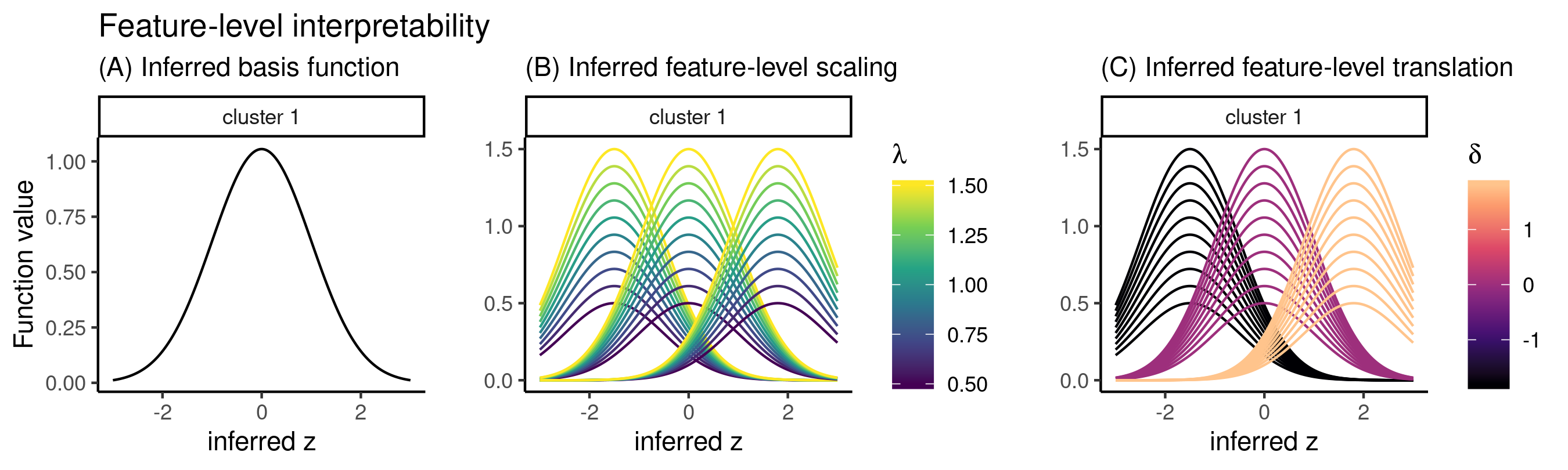}
    \caption{BasisVAE provides feature-level interpretability: Having identified clusters as shown in Fig.~1, not only can we visualise (A) the inferred basis functions, but thanks to the \emph{explicit} parameterisation we have access to the inferred (B) scaling parameters $\lambda$ and (C) translation parameters $\delta$ (shown in colour).}
    \label{fig:basis_parameters}
    \vspace{-5mm}
\end{figure}

High-dimensional data analysis employs dimensionality reduction techniques to extract major components of variation, often using nonlinear techniques, that enable interpretation and visualisation. Additionally, clustering can also be utilised to identify structure both within samples and also across features. 

In this paper, our goal is to develop a general purpose approach for joint dimensionality reduction and clustering of features within the Variational Autoencoder (VAE) framework \citep{kingma_auto-encoding_2014}. Our methodology particularly applies to any \emph{tabular} data problem where individual features have a distinct physical meaning (e.g. a gene, a physiological measurement, the position of a joint on a moving body) and there is an interest in understanding the relationship between these features.  For example, in genomics there is interest in identifying groups of genes with similar behaviour, but the standard VAE framework does provide a mechanism to cluster features. Instead, in practice, often dimensionality reduction and clustering are carried out consecutively in an \textit{ad hoc} manner.

As a motivating example, consider the synthetic gene expression data set shown in Figure \ref{fig:illustration}A. In the heatmap, the columns have been ordered to highlight the existence of five feature groups and the rows arranged to indicate that the observations have a certain ordering. In a real data set this ordering of rows and columns would be \emph{unknown} and part of the statistical inference task would be to identify this \emph{latent} structure, i.e.\ dimensionality reduction. On the other hand, there is also interest in identifying groups of similar features, as shown in Figure \ref{fig:illustration}B. Typically these two tasks are solved separately, whereas we propose a VAE extension to achieve this in a joint probabilistic model. 

Specifically, our approach, which we refer to as BasisVAE, assumes that the features vary as functions of latent dimensions $\boldz$, and that there exists a finite set of such (basis) functions. Hence, BasisVAE aims to discover the scale-invariant clustering of features shown in Figure \ref{fig:illustration}B whilst simultaneously inferring the latent variable $\boldz$ (in this case, $\boldz \in \R$ relates to an ordering of the observations, but in general $\boldz$ may be multi-dimensional). Moreover, in this example, three groups of features have similar shapes which are merely shifted versions of each other. To capture this, we introduce \emph{translation invariance} as part of the model so it would allow these features to be further grouped together, as in Figure \ref{fig:illustration}C, whilst maintaining an explicit parameterisation that would allow the unique properties of each feature to be maintained.

Feature-by-feature similarity measures shown in Figure~\ref{fig:illustration}D help us to further describe our desired output. Under Euclidean or Pearson correlation based distances, the structure of the features is not immediately obvious (note that the features displayed are ordered according to their known groupings which would ordinarily be unknown) whilst k-means clustering ($k=5$) does not uncover this structure either. A linear model is also unsuitable due to the nonlinear relationship between the features and the latent dimensions. In comparison, BasisVAE reveals the correct structure through joint dimensionality reduction and clustering. Translation invariance allows us to further determine whether to treat clusters 1-3 in Figure \ref{fig:illustration}B as three distinct groups of features or as a single group. For easier interpretability, each feature has its own scale and translation parameters as shown in Figure \ref{fig:basis_parameters}.

\begin{figure}[!th]
    \centering
    \includegraphics[width=\columnwidth]{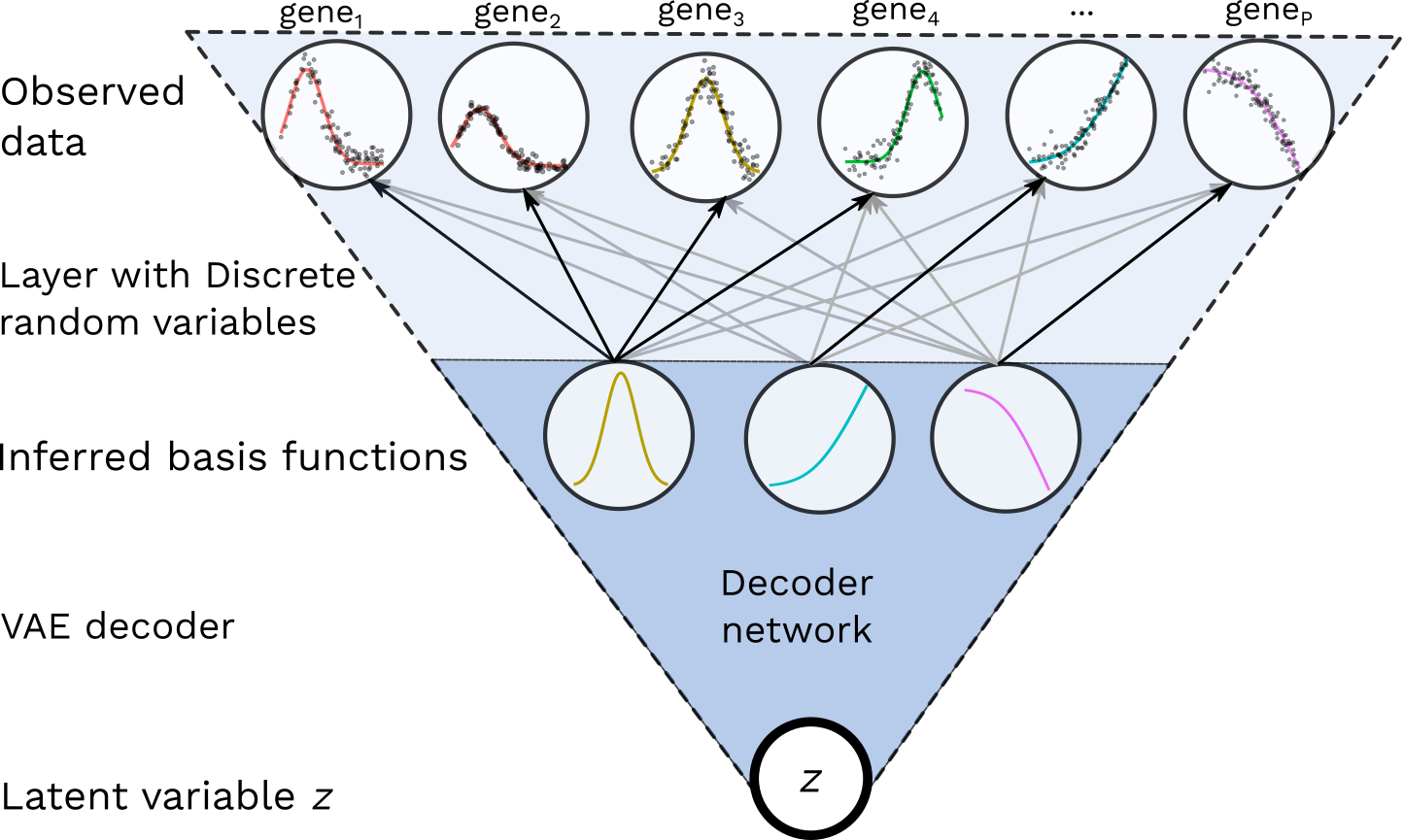}
    \caption{Diagram of the decoder architecture within BasisVAE. From top to bottom: observed data, the inferred basis functions, latent variable $\boldz$. A typical decoder network is used to map $\boldz$ to the basis functions, whereas a specialised layer with Categorical random variables is used to map the latter to features. 
    The posterior distribution of these Categorical variables provides us with the inferred clustering.
    }
    \label{fig:decoder_diagram}
\end{figure}

In this paper, we endow VAEs with the capability to cluster features in the data. 
We propose to achieve this via introducing additional probabilistic structure within the decoder, thus combining the scalability of VAEs with a Bayesian mixture prior for the purpose of clustering features (architecture illustrated in Figure~\ref{fig:decoder_diagram}).

\section{Background: VAEs}


Variational Autoencoders \citep{kingma_auto-encoding_2014} are constructed based on particular a latent variable model structure. Let $\boldy \in \mathbb{R}^P$ denote a data point in some potentially high-dimensional space and $\boldz \in \mathbb{R}^Q$ be a vector of associated latent variables typically in a much lower dimensional space. 
We will assume a prior $\boldz \sim \mathcal{N}(\bf{0}, \bf{I})$. 
Now suppose $f^\theta(\boldz)$ is a family of deterministic functions, indexed by parameters $\theta \in \Theta$, such that $f : \mathbb{R}^Q \times \Theta \rightarrow \mathbb{R}^P$. 
In VAEs, these deterministic functions $f^{\theta}$ are given by deep neural networks (NNs). 
Given data points $\boldy_{1:N} = \{ \boldy_1, \dots, \boldy_N \}$, the marginal likelihood is given by 
$
    p(\boldy_{1:N}) = \prod_{i=1}^N  \int p_\theta(\boldy_i|\boldz_i) p(\boldz_i) d\boldz_i .
$
Here we will assume a Gaussian likelihood, i.e.\ $\boldy_i | \bold z_i, \theta \sim \mathcal{N}(f^\theta(\boldz_i), \sigma^2)$.
Posterior inference in a VAE is carried out via \emph{amortized} variational inference, i.e.\ with a parametric inference model $q_\phi(\boldz_i|\boldy_i)$ where $\phi$ are variational parameters. 
Application of standard variational inference methodology (e.g.\ \citep{blei_variational_2017}) leads to a lower bound on the log marginal likelihood, i.e.\ the ELBO of the form:
\begin{align*}
    \mathcal{L} = \sum_{i=1}^N \mathbb{E}_{q_{\phi}(\boldz_i | \boldy_i)} [\log p_\theta(\boldy_i | \boldz_i)] 
    - \text{KL}(q_{\phi}(\boldz_i|\boldy_i) || p(\boldz_i)) 
\end{align*}

In the VAE, the variational approximation $q_\phi(\boldz_i|\boldy_i)$ is referred to as the \emph{encoder} since it encodes the data $\boldy$ into the latent variable $\boldz$ whilst the \emph{decoder} refers to the generative model $p_\theta(\boldy|\boldz)$ which decodes the latent variables into observations. 

Training a VAE seeks to optimise both the model parameters $\theta$ and the variational parameters $\phi$ jointly using stochastic gradient ascent. This typically requires a stochastic approximation to the gradients of the variational objective which is itself intractable. Typically the approximating family is assumed to be Gaussian, i.e. $q_\phi(\boldz_i|\boldy_i) = \mathcal{N}(\boldz_i|  \mu_{\phi}(\boldy_i), \sigma^2_{\phi}(\boldy_i))$, so that a reparametrisation trick can be used \citep{kingma_auto-encoding_2014, rezende_stochastic_2014}.

We write $f_{\text{decoder}}: \mathbb{R}^Q \to \mathbb{R}^P$ to denote the (multi-output) decoder. But as our interest lies in modelling tabular data where individual features have a distinct meaning, we additionally introduce notation $f_{\text{decoder}}^{(j)}$ where $j$ indexes the $j$-th output dimension of the decoder network (i.e.\ predictions for feature $j$).

\section{BasisVAE}

The goal of our work is to extend the VAE to address the scientific problem of \emph{identifying groups of features} with similar trends. That is, we would like to cluster features. 
We propose to achieve this via a mixture prior as part of the VAE decoding model. The idea is to introduce a set of ``basis'' functions $\{ f_{\text{basis}}^{(k)} : k=1, \ldots, K\}$ parameterised by the neural network, and probabilistically assign every feature $\boldy^{(j)}$ to one of those basis functions $f_{\text{basis}}^{(k)}$. We refer to this as a one-hot-basis functional representation. 

Specifically, we achieve this by replacing the decoder network $f_{\text{decoder}}: \mathbb{R}^Q \to \mathbb{R}^P$ with a basis decoder network $f_{\text{basis}}: \mathbb{R}^Q \to \mathbb{R}^K$ (where the number of basis functions $K < P$) whose output is mapped to the data via Categorical random variables, that is, for every data dimension $j \in \{1, \ldots, P\}$
\begin{equation} \label{eq:basis1}
    f_{\text{decoder}}^{(j)}(\boldz) := \sum_{k=1}^{K} w^{j, k} \cdot f_{\text{basis}}^{(k)}(\boldz)
\end{equation}
where $(w^{j, 1}, \ldots, w^{j, K}) \sim \text{Categorical}(\pi_1, \ldots, \pi_K)$. Note that this can be seen as a mixture model on the \emph{features} rather than observations.
The neural architecture of the BasisVAE decoder has been illustrated in Figure~\ref{fig:decoder_diagram}.

\subsection{Translation and scale invariance}
In the use of basis function representations, a question arises in how to represent two signals which have the same functional shape, but which may be shifted or scaled relative to one another. 
In a pure basis representation (as in equation~\eqref{eq:basis1}), two different basis functions would be learnt, one for each signal, but this may not always be desirable. Sometimes, we would instead like the basis function to only capture the \emph{shape} of the pattern. This can be thought of as an additional layer of clustering in terms of functional similarity across features which is scale- and translation-invariant. 

This behaviour could be achieved via introducing feature-specific scale and translation parameters to \eqref{eq:basis1}. For scale-invariance, we introduce scaling parameters $\lambda_{j, k} \ge 0$ as follows
\begin{equation} \label{eq:basis}
    f_{\text{BasisVAE}}^{(j)}(\boldz) := \sum_{k=1}^{K} w^{j, k} \cdot \lambda_{j, k} \cdot f_{\text{basis}}^{(k)}(\boldz)
\end{equation}
For additional translation-invariance, we introduce $\delta_{j, k}$
\begin{equation} \label{eq:basis_translation_inv}
    f_{\substack{\text{translation-inv} \\ \text{BasisVAE}}}^{(j)}(\boldz) := \sum_{k=1}^{K} w^{j, k} \cdot \lambda_{j, k} \cdot f_{\text{basis}}^{(k)}(\boldz + \delta_{j, k})
\end{equation}
Throughout the paper, we refer to \eqref{eq:basis} as BasisVAE (i.e. we default to scale invariance) and to \eqref{eq:basis_translation_inv} as translation-invariant BasisVAE. 
Note that we have restricted $\lambda_{j, k}$ to be positive. This is because in the genomics applications that are our main interest, we want to distinguish between gene upregulation and downregulation, as these correspond to different biological processes (for this reason, we would not like to merge clusters 4 and 5 in Figure~\ref{fig:illustration}B). 

The added value from scale and translation invariance is two-fold:
\begin{itemize}
    \item Interpretability: Feature-specific scale and/or translation parameters lets us explicitly interpret how features within a cluster relate to each other. 
    \item Increased model capacity: In model given by equation \eqref{eq:basis1}, a large number of basis functions $K$ may be needed to capture the range of observed patterns, whereas \eqref{eq:basis} and \eqref{eq:basis_translation_inv} can provide the same flexibility with a significantly smaller $K$.  
\end{itemize}

\subsection{Specifying the number of basis functions}

So far, we have presented BasisVAE as if the number of underlying basis functions was known. However, this is rarely the case in practice, usually we do not know $K$ \text{a priori}.  Ideally, we would like to be able to over-specify the value of $K$, so that that the unnecessary extra components would be left empty. 

In the mixture model framework outlined above with a $\text{Categorical}(\pi_1, \ldots, \pi_K)$ prior, this would require careful specification of $\boldpi = (\pi_1, \ldots, \pi_K)$ which would be problem dependent. Hierarchical Bayesian models offer a way to get around this by placing a distribution over $\boldpi$. One possible construction would be as follows  
\begin{align} \label{eq:DirCat}
    \begin{split}
    \boldpi | \boldalpha &\sim \Dir(\boldalpha) \\
    (w^{j, 1}, \ldots, w^{j, K}) | \boldpi &\sim \Cat(\pi_1, \ldots, \pi_K)
    \end{split}
\end{align}
This hierarchical construction can be useful, because for $\alpha < 1$ values the draws from $p(\boldpi | \boldalpha)$ will be sparse, thus providing a mechanism to \emph{learn} an appropriately sparse $\boldpi$ rather than pre-specify its value. This particular setup has been popular for finite Dirichlet mixture models as well as their Dirichlet Process counterparts, where conjugacy made it possible to develop efficient (collapsed) Gibbs samplers, see e.g.\ \citep{neal_markov_2000}. MCMC-based inference has perhaps been most popular for such mixture models, but also variational inference techniques have been derived (e.g.\ analytic update rules for exponential families \citep{blei_variational_2006}).

However, for non-linear latent variable models, such as VAEs, analytic updates are not available in closed form any more. Also it is not straightforward to exploit conjugacy in eq. \eqref{eq:DirCat}. This imposes a challenge: on the one hand we want to utilise the computational benefits (scalability, modularity, computational efficiency) of the VAE-framework, while on the other hand we would like to make use of the sparsity properties that are present in more classical (``pre automatic differentiation era'') hierarchical Bayesian mixture models. 

Existing work in the context of VAEs\footnote{To our knowledge, existing work has only considered mixtures for clustering observations rather than features} have either relied on using a fixed value of $\boldpi = (1/K, \ldots, 1/K)$ \citep{dilokthanakul_deep_2016} or developed inference schemes where variational approximation is introduced on the joint space of $(\boldpi, \boldw)$ assuming factorisation $q(\boldw, \boldpi) = q(\boldw) q(\boldpi)$ \citep{nalisnick_approximate_2016}. 
The former approach is undesirable, as it has no mechanism to encourage sparsity. The latter could in principle work well, but its performance could suffer due to ignoring the (potentially strong) dependence between cluster allocations $\boldw$ and the $\boldpi$. 

Next, we show how our attempts to adapt these ideas 
lead to \emph{inefficient} inference schemes, before proposing a modern adaptation of the \emph{collapsed} inference scheme within the VAE framework. 

\subsection{Generative model and inference}

The generative model for the scale-and-translation-invariant BasisVAE is as follows
\begin{align*}
    \boldz_i &\sim \mathcal{N}(\boldzero, \boldI) \\
    \boldpi | \boldalpha &\sim \Dir(\boldalpha) \\
    (w^{j, 1}, \ldots, w^{j, K}) | \boldpi &\sim \Cat(\pi_1, \ldots, \pi_K) \\
    y_i^{(j)} | \boldw^j, \boldz_i, \theta, \lambda, \delta &\sim \mathcal{N}\left(\sum_{k=1}^{K} w^{j, k} \lambda_{j,k} f_{\text{basis}}^{(k)}(\boldz_i + \delta_{jk}), \sigma^2_j \right)
\end{align*}
where $i=1, \ldots, N$ indexes data points, $j=1, \ldots,P$ indexes features, and $k=1, \ldots, K$ basis components. 
We additionally place a prior over $\delta_{jk} \sim \N(0, 1)$ and $\lambda_{jk} \sim \Gamma(1, 1)$, but in our experiments we will perform MAP estimation over $\delta_{jk}$ and $\lambda_{jk}$. 

Denoting observations collectively $\boldY := \{\boldy_i^{(j)}\}$, 
standard VAE methodology would lower bound the log marginal likelihood as follows $\log p(\boldY) \ge$ 
\begin{align*}
    \ge 
    \sum_{i=1}^N
    \mathbb{E}_{q_{\phi}(\boldz_i | \boldy_i)} [\log p_\theta(\boldy_i | \boldz_i)] - \text{KL}(q_{\phi}(\boldz_i|\boldy_i) || p(\boldz_i))
\end{align*}
However, unlike for VAE, in our model $\log p_\theta(\boldy_i | \boldz_i)$ is intractable due to the need to integrate over $\boldw$ and $\boldpi$. 

\subsubsection{Non-collapsed inference}

\setlength{\abovedisplayskip}{3pt}
\setlength{\belowdisplayskip}{3pt}

One approach would be to introduce an approximate posterior $q(\boldw, \boldpi) = \prod_{j=1}^P q(\boldw^j) q(\boldpi)$. For example, one can choose variational families $q(\boldw^j) = \Cat(\phi_{j, 1}, \ldots, \phi_{j, K})$ and $q(\boldpi)=\Dir(\boldsymbol{\psi})$ where $\boldsymbol{\phi}$ and $\boldsymbol{\psi}$ are variational parameters. We note that reparameterisation trick for the Dirichlet can be implemented as in \citep{figurnov_implicit_2018}. 
Assuming mean-field across $\{\boldz_i\}, \{\boldw^j\}, \boldpi$, we obtain the ELBO
\begin{align*}
    &\sum_{i=1}^N
    \mathbb{E}_{q_{\phi}(\boldz_i | \boldy_i)} \mathbb{E}_{q(\boldw)} \log p_\theta(\boldy_i | \boldz_i, \boldw) 
    - \text{KL}(q_{\phi}(\boldz_i|\boldy_i) || p(\boldz_i)) \\
    & - \text{KL} \left( \prod_{j=1}^P q(\boldw^j) q(\boldpi) \; || \; \prod_{j=1}^P p(\boldw^j | \boldpi) p(\boldpi) \right)
\end{align*}
We refer to this approach as \emph{non-collapsed} inference, and later demonstrate how this inference scheme can get stuck in local modes, resulting in undesirable behaviour.

\subsubsection{Collapsed inference}

Given the success of earlier work on collapsed inference schemes, e.g. collapsed Gibbs samplers \citep{neal_markov_2000} or collapsed VB schemes \citep{teh_collapsed_2007, kurihara_collapsed_2007, hensman_fast_2012, hensman_fast_2015}, we hope to improve our previously decribed non-collapsed inference by marginalising out $\boldpi$. 
If we adapted the approach of \citet{teh_collapsed_2007}, further approximations for evaluating the ELBO would be needed. Instead, we follow the approach of \citet{hensman_fast_2012, hensman_fast_2015} where marginalisation is applied \emph{after} (rather than before) making a variational approximation. 

Now, following this strategy, we lower bound the term $\log p_\theta(\boldy | \boldz)$, by first introducing an approximate posterior
\begin{equation*}
    q(\boldw) = \prod_{j=1}^P q(\boldw^j) = \prod_{j=1}^P \Cat(\phi_{j, 1}, \ldots, \phi_{j, K})
\end{equation*}
and then collapsing $\boldpi$ from the bound without any further approximations. The detailed derivation can be found in Supplementary~\ref{supp:ELBO}. 
As a result, we obtain the ELBO
\begin{align}
    \calL = & 
    \sum_{i=1}^N
    \mathbb{E}_{q_{\phi}(\boldz_i | \boldy_i)} \mathbb{E}_{q(\boldw)} \log p_\theta(\boldy_i | \boldz_i, \boldw) \label{eq:ELBO1} \\
    &+ \log \int \exp \left( \mathbb{E}_{q(\boldw)} p(\boldw | \boldpi) \right) p(\boldpi) d\pi \label{eq:ELBO2} \\
    &- \mathbb{E}_{q(\boldw)} \log q(\boldw) \label{eq:ELBO3} \\
    &- \sum_{i=1}^N \text{KL}(q_{\phi}(\boldz_i|\boldy_i) || p(\boldz_i)) \label{eq:ELBO4}
\end{align}
For the first term~\eqref{eq:ELBO1}, we use a combination of reparameterisation and analytic marginalisation, the former for sampling $\boldz_i \sim q_{\phi}(\boldz_i | \boldy_i)$ and the latter for $q(\boldw)$. So we evaluate
\begin{align*}
    \sum_{i=1}^N \mathbb{E}_{q(\boldz_i | \boldy_i)} \sum_{j=1}^P \sum_{k=1}^K \phi_{j,k} \log \mathcal{N}(y_i^{(j)} | \lambda_{j,k} f_{\text{basis}}^{(k)}(\boldz_i + \delta_{jk}), \; \sigma^2_j) .
\end{align*}
The second term~\eqref{eq:ELBO2} can be derived analytically, 
\begin{align*}
    \log \frac{\Gamma(\sum_k \alpha_k)}{\prod_k \Gamma(\alpha_k)} \int \prod_{k=1}^K  \pi_k^{n_k + \alpha_k - 1} d \boldpi
\end{align*}
where $n_k := \sum_{j=1}^P \phi_{j, k}$. This integral has a closed form, resulting in the analytic term 
\begin{equation*}
    \log \frac{\Gamma(\sum_k \alpha_k)}{\prod_k \Gamma(\alpha_k)}
\frac{\prod_k \Gamma(n_k + \alpha_k)}{\Gamma(\sum_k n_k + \alpha_k)} .
\end{equation*}
Also the third and fourth term have analytic forms. 

Finally, we perform MAP inference over $\lambda_{jk}$ and $\delta_{jk}$ (but we note that variational inference could also be adapted), encouraging $\delta_{jk}$ values to be centered around 0 and $\lambda_{jk}$ values centered around 1. So the final optimisation objective will incorporate additional penalties
\begin{equation*}
    \calL + \sum_{j=1}^P \sum_{k=1}^K \log \N(\delta_{jk} \,|\, 0, 1) + \log \Gamma(\lambda_{jk} \,|\, 1, 1) .
\end{equation*}

\textbf{Adaptation of ELBO for large data sets:}
For large high-dimensional data sets, we note the lower bound $\calL$ will be dominated by the data log-likelihood term~\eqref{eq:ELBO1}. As a result, our clustering prior will implicitly become less important for increasingly large $N$ and $P$. While the property that the likelihood will dominate the prior in the large data regime is inherent to Bayesian models, it may not always be desirable, especially for mis-specified models. In practice, this can be alleviated via downweighing the likelihood or upweighing the prior.
For example, $\beta$-VAE \citep{higgins_beta-vae:_2017} scales the KL-term by $\beta>0$, which is closely connected to the ELBO on an alternative formulation involving an annealed prior $\propto p(\boldz)^\beta$ \citep{hoffman_beta-vaes_2017, mathieu_disentangling_2019}.
Analogously to the $\beta$-VAE, we propose a modified objective $\calL_\beta$ (as outlined in Supplementary~\ref{supp:beta}) where $\beta > 1$ lets us increase the relative importance of the sparse clustering prior for large-scale applications. 

\textbf{Scalability:} As a result, we have derived an inference scheme which can be easily implemented as a VAE extension, thus making it possible to leverage the advantages of deep learning frameworks. From the computational efficiency perspective, the ELBO for the standard VAE involves summing over (a subset of) observations and all features, whereas our ELBO includes an additional sum over cluster components $k=1, \ldots, K$.  Thus the additional scaling w.r.t.\ $K$ is linear (but we note that this can be efficiently implemented by introducing an extra dimension to tensors). 

\section{Related work}

To our knowledge, the problem of \emph{clustering features} has not been previously addressed in the context of VAEs. 
In the regression setting, various probabilistic methods for curve clustering have been proposed before.  E.g.\ 
\citet{chudova_translation-invariant_2003} propose a method for translation-invariant curve clustering, assuming fixed inputs on an equi-spaced time grid. \citet{gaffney_joint_2005} derive an EM-algorithm for curve clustering and alignment, assuming that curves are parameterised using either polynomial regression or cubic splines. 
\citet{palla_nonparametric_2012} propose a Bayesian non-parametric \emph{linear} latent variable model. Thus, it could be seen as a special case of BasisVAE with a \emph{linear} decoder and infinite $K$. However, the linearity assumption is too restrictive for the applications we consider in this paper. 
\citet{campbell_descriptive_2018} propose a specific translation-invariant framework for single-cell genomics, however it captures only two types of parametric mappings: sigmoidal and transient patterns. This is much more restrictive compared to BasisVAE that can \emph{learn} the dynamics in an unsupervised way.

\section{Experiments}

It is challenging to evaluate the performance of clustering methods on real applications, as the ground truth cluster labels are rarely known (and even if labels exist, there may always be more fine-grained subgroups). For this reason, we first evaluate the performance of BasisVAE in a controlled setting, before considering the semi-synthetic clustering of motion capture data, and finally demonstrating the utility of our approach for the analysis of single-cell gene expression data. 

Our PyTorch implementation of BasisVAE is available in \url{https://github.com/kasparmartens/BasisVAE}.

\subsection{Toy data}

\begin{figure}[!th]
    \centering
    \includegraphics[width=\columnwidth]{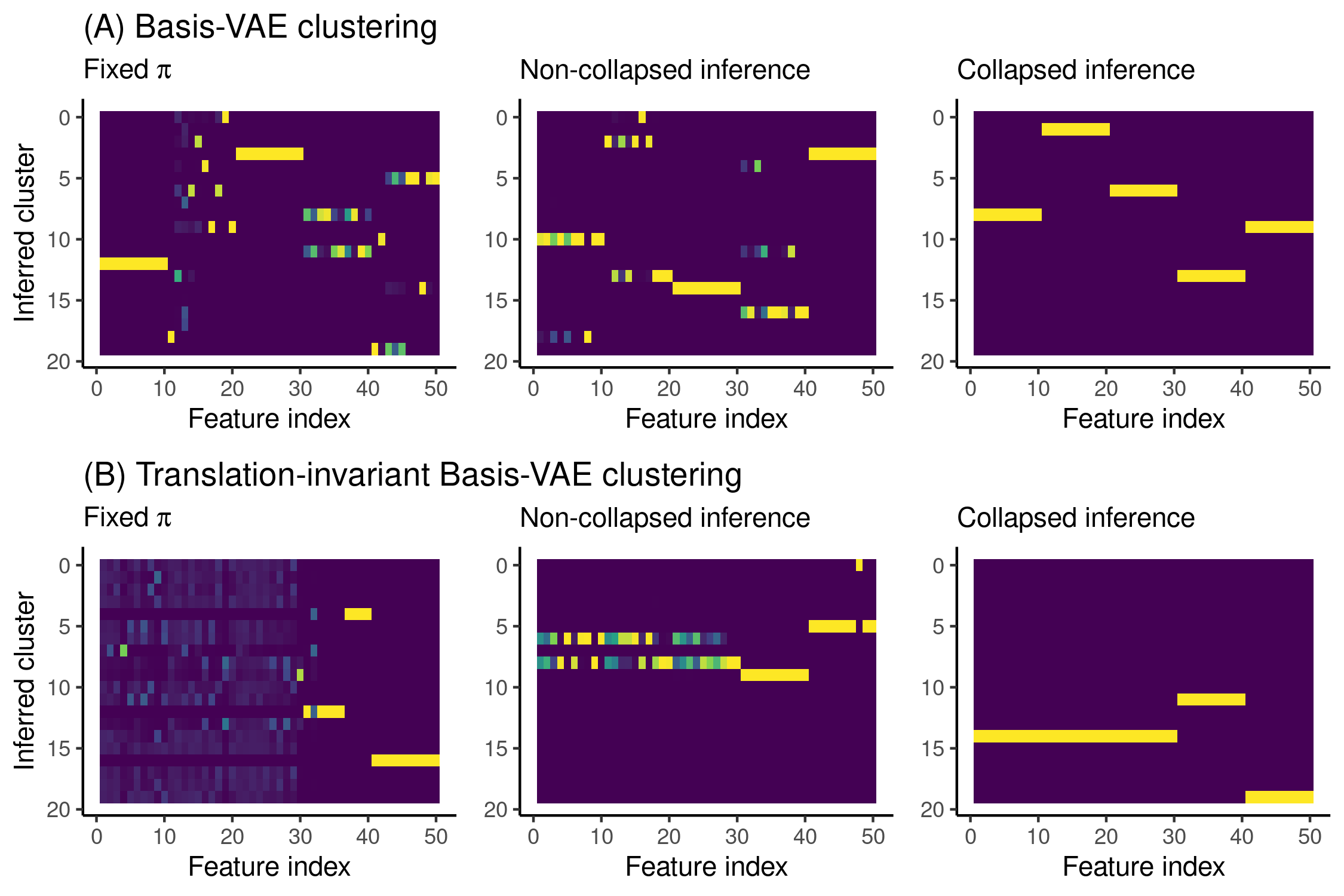}
    \caption{Comparison of inference techniques for $q(\boldw, \boldpi)$, for both (A) BasisVAE and (B) its translation-invariant version, using the synthetic data as in Fig~\ref{fig:illustration} (consisting of 5 groups with 10 features each). The heatmaps show posterior probabilities for cluster allocations, i.e. elements of $q(\boldw)$. Both ``fixed $\boldpi$'' and ``non-collapsed'' inference (left and middle panels) result in suboptimal and non-sparse cluster assignments, whereas collapsed inference correctly identifies the underlying ((A) $K=5$ and (B) $K=3$) clusters.}
    \label{fig:toy_collapsed}
    \vspace{3mm}
    \centering
    \includegraphics[width=\columnwidth]{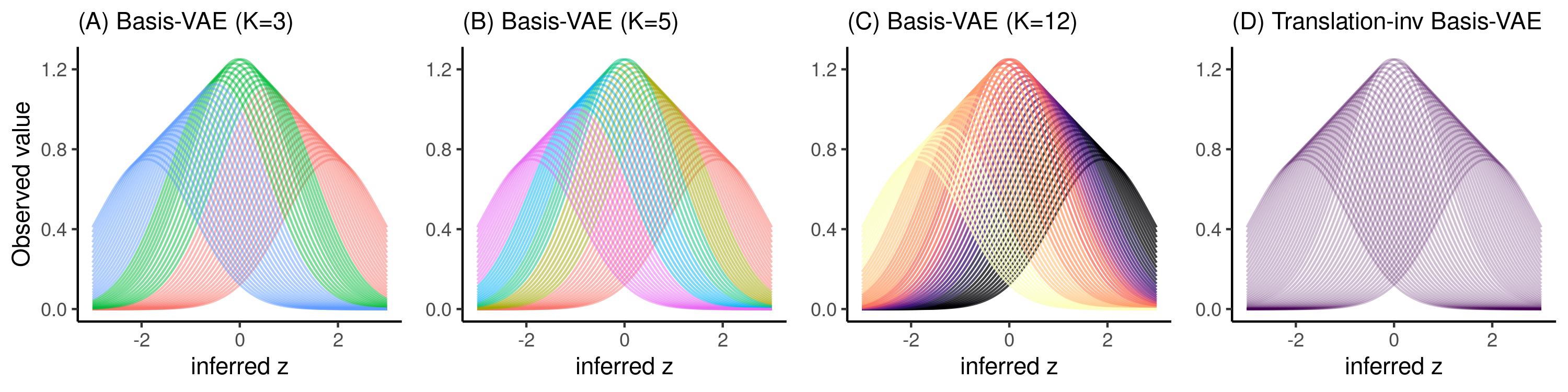}
    \caption{
        A large number of basis functions is required to accurately capture the set of shifted mappings (cluster allocations shown in colour for BasisVAE using (A) $K$=3, (B) $K$=5, and (C) $K$=12), whereas (D) translation-invariant BasisVAE can represent this with a single basis function. 
    }
    \label{fig:toy2}
    \vspace{-3mm}
\end{figure}

We first consider the synthetic data shown in Figure~\ref{fig:illustration} with 5 distinct groups of features (with 10 features each, $P=5 \times 10$) as a function of latent $\boldz \in \mathbb{R}$. The data generative process was designed in order to mimic the patterns we expect to see in real gene expression data: first three groups of features exhibit a transient trend (where gene expression spikes for a certain range of $\boldz$ values), whereas the fourth cluster exhibits gene up-regulation and fifth one down-regulation. The BasisVAE correctly infers five clusters (Figure~\ref{fig:illustration}B), whereas its translation-invariant version has merged the first three clusters together (Figure~\ref{fig:illustration}C).

Note that these results were obtained under our collapsed inference scheme. This would not have been possible with simpler approaches such as the ``fixed $\boldpi$'' and ``non-collapsed'' inference, as shown in Figure~\ref{fig:toy_collapsed}, which result in cluster allocations that are less sparse than the one inferred by ``collapsed'' inference (all comparisons were carried out with $K$ fixed to $K=20$ and $\alpha=0.1$). ``Fixed $\boldpi$'' has failed likely due to the reason that $\pi_k = 1/K$ does not encourage sparse solutions, whereas ``non-collapsed'' inference is more prone to getting stuck in local modes due to the dependence between $\boldw$ and $\boldpi$. Clustering performance has been quantified in Supp~\ref{supp:fig_synthetic} using the V-measure \citep{rosenberg_v-measure:_2007} over 10 random restarts.

Various widely used methods for clustering gene expression data rely on the feature-by-feature similarity matrix. As shown in Fig.~\ref{fig:illustration}D, these do not accurately capture the underlying cluster structure, and do not provide a way to naturally incorporate scale and translation invariance. Also the linear BasisVAE, which can be seen as a particular implementation of \citep{palla_nonparametric_2012}, does not provide the correct clustering either.

Next, we demonstrate the capabilities of BasisVAE to capture translation invariance in a setting where we observe features spanning a range of $\delta$ values (Fig.~\ref{fig:toy2}). The non-translation-invariant BasisVAE clustering behaves intuitively: for small $K \in \{3, 5, 12\}$ (panels A-C), features with similar $\delta$ values are clustered together (cluster allocations shown in colour). This may be sufficient for certain applications, but there is no way to quantify how different clusters relate to one another. This capability is present in the translation-invariant BasisVAE, which identifies a single cluster (Fig~\ref{fig:toy2}D) and relates all features within the cluster to one another explicitly via inferred $\delta$.

\subsection{Motion capture data}

Next we investigate the behaviour of our model in a semi-controlled setting. We use CMU Motion Capture database which consists of short segments of various activities over a repeated number of trials. Note that even if we focus on the same activity performed by the same subject, the measurements (the locations of joints) are not \emph{synchronised} between repeated trials. We note that this alignment problem has a particular structure which is inherently simpler than the structure in the genomics data in the next section. First, unlike for the genomics application, for motion capture data time is observed. Second, we know that we need to align measurements only \emph{between} trials, but not within a trial (if measurements for one joint are shifted by $\delta$, it is likely the case for other joints as well). 

Here, we focus on the 10 repeated trials of the activity ``golf swing''. 
For the purpose of demonstrating the capabilities of BasisVAE, we treat 10 trials of 59-dimensional measurements as a high-dimensional data set with $P=590$ dimensions without \textit{a priori} known structure. Our goal is three-fold: First, we hope to uncover clusters that are shared \emph{across} trials, e.g.\ we expect the ``right hand'' movements to follow the same dynamics over all trials. Second, we hope to explicitly align the movements across trials. Third, we would like to find \emph{shared} dynamics between the marker positions, e.g.\ two nearby joints may exhibit similar movement. 

\begin{figure}[!th]
    \centering
    \includegraphics[width=\columnwidth]{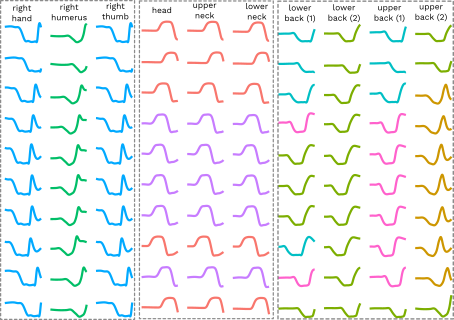}
    \caption{BasisVAE on motion capture data. The inferred mappings together with the clustering (shown in colour), shown over 10 repeated trials (in rows) and selected features (in columns). Features have been retrospectively grouped into 1) right-hand, 2) head-and-neck, and 3) lower-and-upper-back related features.}
    \label{fig:mocap}
    \vspace{3mm}
    \centering
    \includegraphics[width=\columnwidth]{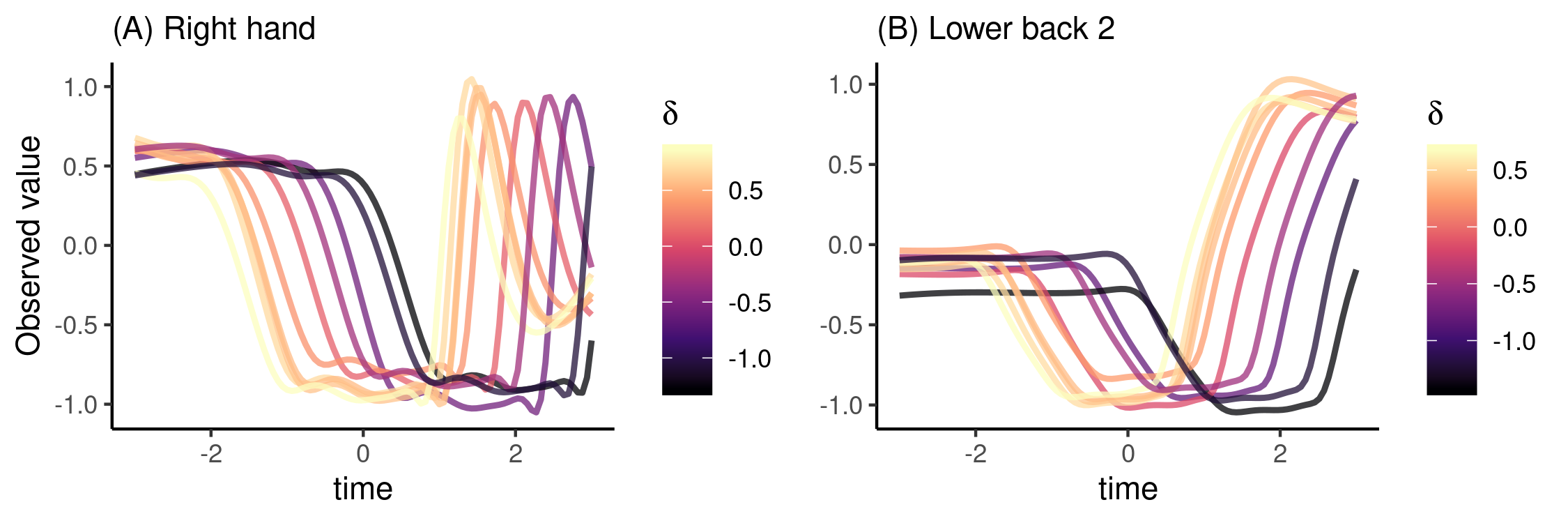}
    \caption{BasisVAE lets us explicitly align features in motion capture data, demonstrated on features ``right hand'' and ``lower back (2)'' ($\delta$ values shown in colour).}
    \label{fig:mocap_delta}
    \vspace{-5mm}
\end{figure}

We have visualised results in Fig~\ref{fig:mocap} for a subset of features (in columns) and all 10 trials (in rows), showing the inferred mappings (lines) and the respective cluster allocations (in colour). The features can be grouped into 1) right-hand , 2) head-and-neck, and 3) lower-and-upper-back related movement. Note that for many features, all 10 trials have been assigned to the same cluster, whereas sometimes there are subtle differences between the trials (e.g.\ for head and neck, the ``bump'' for some trials lasts slightly longer than for others). 
Our model provides interpretability that is not present in alternative methods. We can align features between trials (inferred $\delta$ values shown in Figure~\ref{fig:mocap_delta}), and identify groups of similar features. E.g.\ ``right hand'' and ``right thumb'' have been assigned the same cluster, whereas ``right humerus'' exhibits a different dynamic.

\subsection{Single-cell Spermatogenesis data}

\begin{figure*}[!th]
    \centering
    \includegraphics[width=\textwidth]{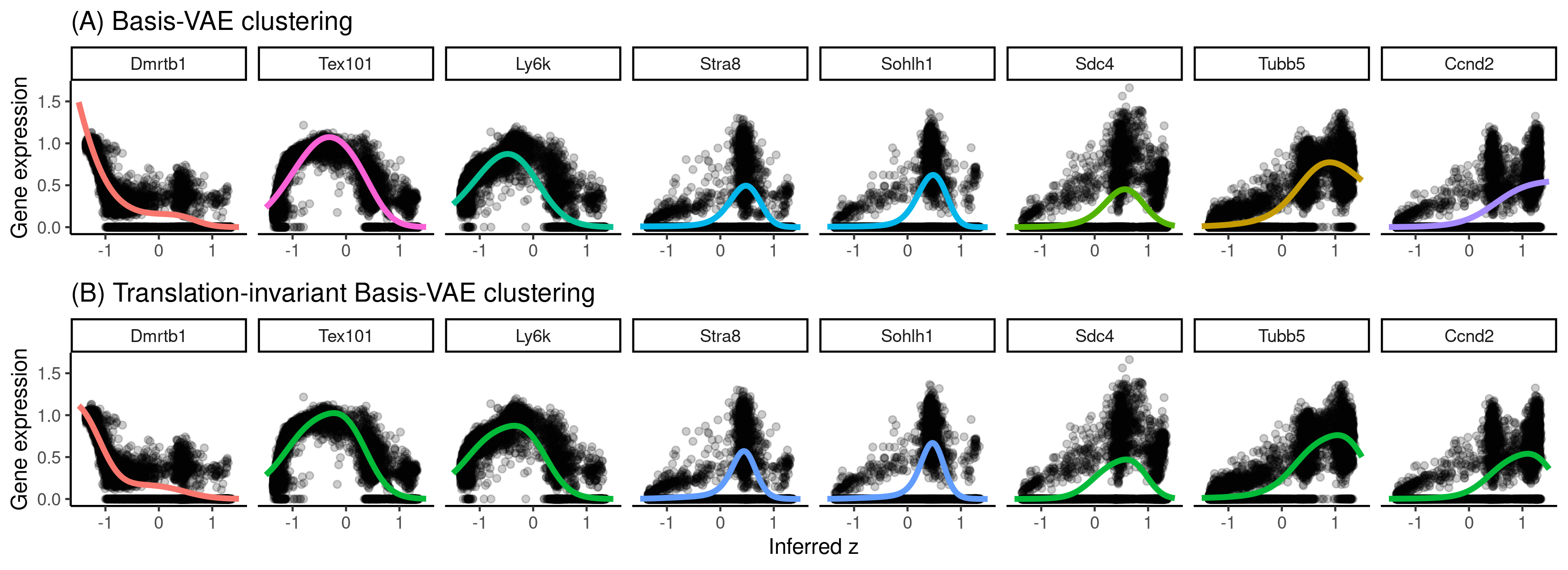}
    \vspace{-8mm}
    \caption{BasisVAE has discovered structure in the large-scale single-cell spermatogenesis data. For selected genes (whose peak expression along the $z$-axis varies from left to right, i.e.\ from Dmrtb1 to Ccnd2), we have shown how (A) BasisVAE has identified a number of clusters with distinct gene expression behaviour (clusters shown in colour), whereas (B) its translation-invariant version has grouped genes with similar profiles together (clusters shown in colour), thus aiding further interpretability. }
    \label{fig:spermatogenesis}
    \vspace{-5mm}
\end{figure*}

\begin{figure}
    \centering
    \includegraphics[width=\columnwidth]{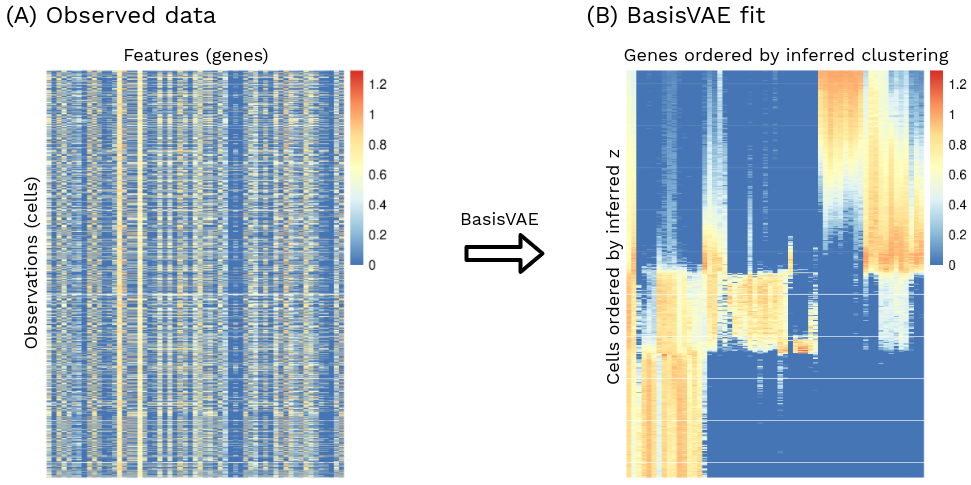}
    \vspace{-4mm}
    \caption{BasisVAE has uncovered structure in spermatogenesis data. (A) Heatmap of the observed data for a subset of genes. (B) Same data with re-ordered rows (ordered by $z$) and columns (ordered by clusters).}
    \label{fig:spermatogenesis_heatmap}
    \vspace{-6mm}
\end{figure}

Recently, there has been a lot of interest in single-cell genomics, where gene expression is measured on the level of individual cells. This results in large data sets, both in the number of samples (i.e.\ cells) as well as the number of features (i.e.\ genes). The ability to quantify individual cells has led to a number of studies where the aim is to characterise \emph{cellular differentiation} -- the process by which a cell changes from one cell type to another. As the technology lets us measure every cell only once, we do not have access to the true underlying time, how far along the differentiation trajectory every cell is. Those quantities are referred to as \emph{pseudotimes} \citep{trapnell_dynamics_2014}, and are treated as one-dimensional latent variables $\boldz$. 

The main scientific interest in these studies lies in getting insights into the biological processes that drive this latent dimension, e.g.\ identifying groups of genes whose expression changes with $\boldz$, and characterising those changes. 
In this section, we demonstrate the utility of BasisVAE for such purposes. Here, we re-analyse a publicly available data set from a recently published study \cite{ernst_staged_2019}, where the authors used single-cell RNA-Sequencing to profile mouse spermatogenesis. 
Due to the data being non-negative-valued and exhibiting zero-inflation, here we follow a common choice (e.g.\ as in \citep{lopez_deep_2018}) and we use the zero-inflated negative binomial likelihood in the decoder. 


We pick a subset of samples ($N=8509$) and $P=5000$ most variable genes, and fit BasisVAE without and with translation invariance (we used $K=50$ clusters). 
Supp~\ref{supp:fig_singlecell} illustrates how the inferred $\boldz$ relates to the first two principal components (PC1, PC2) of the data. 
Manually interpreting individual genes in such a large-scale setting is infeasible, however the inferred BasisVAE cluster allocations make interpretation easier. In Figure~\ref{fig:spermatogenesis} we display some of the genes that were highlighted in the analysis of \citet{ernst_staged_2019} together with our inferred mappings and cluster allocations. Note that in the scale-invariant version, genes exhibiting the same shape but different shift were allocated different clusters, whereas e.g.\ genes \textit{Stra8} and \textit{Sohlh1} were identified as exhibiting the same behaviour. This is different from the translation-invariant model, where e.g.\ genes \textit{Tex101, Ly6k, Sdc4, Tubb5, Ccnd2} were all assigned to the same cluster, so not only do we know that they have the same shape, but we can also interpret their relative shift by inspecting the 
$\delta$ values. 

While the inferred clustering structure (e.g.\ where the behaviour of gene might be delayed relative to another, but exhibiting the same shape) does not imply any biological functional relatedness, BasisVAE methodology can add to the exploratory toolkit for investigators to explore and rank such potential relationships between genes explicitly rather than retrospectively through \textit{ad hoc} approaches that are common in genomics.

\section{Discussion}

We have proposed a joint model for dimensionality reduction and clustering of features, implemented in the VAE framework. Specifically, we have modified the decoder to incorporate a hierarchical Bayesian clustering prior, and demonstrated how collapsed variational inference can identify sparse solutions when over-specifying $K$. 
Furthermore, translation invariance lets us handle scenarios where not all features are \emph{aligned}.

\subsection*{Acknowledgements}

KM was supported by a UK Engineering and Physical Sciences Research Council Doctoral Studentship. CY is supported by a UK Medical Research Council Research Grant (Ref: MR/P02646X/1) and by The Alan Turing Institute under the EPSRC grant EP/N510129/1.

\bibliography{references}

\begin{thebibliography}{}

\bibitem[Blei and Jordan, 2006]{blei_variational_2006}
Blei, D.~M. and Jordan, M.~I. (2006).
\newblock Variational inference for {Dirichlet} process mixtures.
\newblock {\em Bayesian analysis}, 1(1):121--143.

\bibitem[Blei et~al., 2017]{blei_variational_2017}
Blei, D.~M., Kucukelbir, A., and McAuliffe, J.~D. (2017).
\newblock Variational inference: {A} review for statisticians.
\newblock {\em Journal of the American Statistical Association},
  112(518):859--877.

\bibitem[Campbell and Yau, 2018]{campbell_descriptive_2018}
Campbell, K.~R. and Yau, C. (2018).
\newblock A descriptive marker gene approach to single-cell pseudotime
  inference.
\newblock {\em Bioinformatics}, 35(1):28--35.

\bibitem[Chudova et~al., 2003]{chudova_translation-invariant_2003}
Chudova, D., Gaffney, S., Mjolsness, E., and Smyth, P. (2003).
\newblock Translation-invariant mixture models for curve clustering.
\newblock In {\em Proceedings of the ninth {ACM} {SIGKDD} international
  conference on {Knowledge} discovery and data mining}, pages 79--88. ACM.

\bibitem[Dilokthanakul et~al., 2016]{dilokthanakul_deep_2016}
Dilokthanakul, N., Mediano, P.~A., Garnelo, M., Lee, M.~C., Salimbeni, H.,
  Arulkumaran, K., and Shanahan, M. (2016).
\newblock Deep unsupervised clustering with gaussian mixture variational
  autoencoders.
\newblock {\em arXiv preprint arXiv:1611.02648}.

\bibitem[Ernst et~al., 2019]{ernst_staged_2019}
Ernst, C., Eling, N., Martinez-Jimenez, C.~P., Marioni, J.~C., and Odom, D.~T.
  (2019).
\newblock Staged developmental mapping and {X} chromosome transcriptional
  dynamics during mouse spermatogenesis.
\newblock {\em Nature communications}, 10(1):1251.

\bibitem[Figurnov et~al., 2018]{figurnov_implicit_2018}
Figurnov, M., Mohamed, S., and Mnih, A. (2018).
\newblock Implicit {Reparameterization} {Gradients}.
\newblock In Bengio, S., Wallach, H., Larochelle, H., Grauman, K.,
  Cesa-Bianchi, N., and Garnett, R., editors, {\em Advances in {Neural}
  {Information} {Processing} {Systems} 31}, pages 441--452. Curran Associates,
  Inc.

\bibitem[Gaffney and Smyth, 2005]{gaffney_joint_2005}
Gaffney, S.~J. and Smyth, P. (2005).
\newblock Joint probabilistic curve clustering and alignment.
\newblock In {\em Advances in neural information processing systems}, pages
  473--480.

\bibitem[Hensman et~al., 2012]{hensman_fast_2012}
Hensman, J., Rattray, M., and Lawrence, N.~D. (2012).
\newblock Fast variational inference in the conjugate exponential family.
\newblock In {\em Advances in neural information processing systems}, pages
  2888--2896.

\bibitem[Hensman et~al., 2015]{hensman_fast_2015}
Hensman, J., Rattray, M., and Lawrence, N.~D. (2015).
\newblock Fast {Nonparametric} {Clustering} of {Structured} {Time}-{Series}.
\newblock {\em IEEE Transactions on Pattern Analysis and Machine Intelligence},
  37(2):383--393.

\bibitem[Higgins et~al., 2017]{higgins_beta-vae:_2017}
Higgins, I., Matthey, L., Pal, A., Burgess, C., Glorot, X., Botvinick, M.,
  Mohamed, S., and Lerchner, A. (2017).
\newblock beta-{VAE}: {Learning} {Basic} {Visual} {Concepts} with a
  {Constrained} {Variational} {Framework}.
\newblock {\em International Conference on Learning Representations}.

\bibitem[Hoffman et~al., 2017]{hoffman_beta-vaes_2017}
Hoffman, M.~D., Riquelme, C., and Johnson, M.~J. (2017).
\newblock The beta-{VAE}’s {Implicit} {Prior}.
\newblock In {\em Workshop on {Bayesian} {Deep} {Learning}, {NIPS}}, pages
  1--5.

\bibitem[Kingma and Welling, 2014]{kingma_auto-encoding_2014}
Kingma, D.~P. and Welling, M. (2014).
\newblock Auto-encoding variational bayes.
\newblock {\em Proceedings of the International Conference on Learning
  Representations (ICLR)}.

\bibitem[Kurihara et~al., 2007]{kurihara_collapsed_2007}
Kurihara, K., Welling, M., and Teh, Y.~W. (2007).
\newblock Collapsed {Variational} {Dirichlet} {Process} {Mixture} {Models}.
\newblock In {\em {IJCAI}}, volume~7, pages 2796--2801.

\bibitem[Lopez et~al., 2018]{lopez_deep_2018}
Lopez, R., Regier, J., Cole, M.~B., Jordan, M.~I., and Yosef, N. (2018).
\newblock Deep generative modeling for single-cell transcriptomics.
\newblock {\em Nature methods}, 15(12):1053.

\bibitem[Mathieu et~al., 2019]{mathieu_disentangling_2019}
Mathieu, E., Rainforth, T., Siddharth, N., and Teh, Y.~W. (2019).
\newblock Disentangling {Disentanglement} in {Variational} {Autoencoders}.
\newblock In {\em International {Conference} on {Machine} {Learning}}, pages
  4402--4412.

\bibitem[Nalisnick et~al., 2016]{nalisnick_approximate_2016}
Nalisnick, E., Hertel, L., and Smyth, P. (2016).
\newblock Approximate inference for deep latent gaussian mixtures.
\newblock In {\em {NIPS} {Workshop} on {Bayesian} {Deep} {Learning}}.

\bibitem[Neal, 2000]{neal_markov_2000}
Neal, R.~M. (2000).
\newblock Markov {Chain} {Sampling} {Methods} for {Dirichlet} {Process}
  {Mixture} {Models}.
\newblock {\em Journal of Computational and Graphical Statistics},
  9(2):249--265.

\bibitem[Palla et~al., 2012]{palla_nonparametric_2012}
Palla, K., Ghahramani, Z., and Knowles, D.~A. (2012).
\newblock A nonparametric variable clustering model.
\newblock In {\em Advances in {Neural} {Information} {Processing} {Systems}},
  pages 2987--2995.

\bibitem[Rezende et~al., 2014]{rezende_stochastic_2014}
Rezende, D.~J., Mohamed, S., and Wierstra, D. (2014).
\newblock Stochastic backpropagation and approximate inference in deep
  generative models.
\newblock {\em arXiv preprint arXiv:1401.4082}.

\bibitem[Rosenberg and Hirschberg, 2007]{rosenberg_v-measure:_2007}
Rosenberg, A. and Hirschberg, J. (2007).
\newblock V-measure: {A} conditional entropy-based external cluster evaluation
  measure.
\newblock In {\em Proceedings of the 2007 joint conference on empirical methods
  in natural language processing and computational natural language learning
  ({EMNLP}-{CoNLL})}, pages 410--420.

\bibitem[Teh et~al., 2007]{teh_collapsed_2007}
Teh, Y.~W., Newman, D., and Welling, M. (2007).
\newblock A collapsed variational {Bayesian} inference algorithm for latent
  {Dirichlet} allocation.
\newblock In {\em Advances in neural information processing systems}, pages
  1353--1360.

\bibitem[Trapnell et~al., 2014]{trapnell_dynamics_2014}
Trapnell, C., Cacchiarelli, D., Grimsby, J., Pokharel, P., Li, S., Morse, M.,
  Lennon, N.~J., Livak, K.~J., Mikkelsen, T.~S., and Rinn, J.~L. (2014).
\newblock The dynamics and regulators of cell fate decisions are revealed by
  pseudotemporal ordering of single cells.
\newblock {\em Nature biotechnology}, 32(4):381.

\end{thebibliography}

\clearpage

\onecolumn

\begin{appendices}

\section*{Supplementary Information}

\section{Derivation of the collapsed ELBO} \label{supp:ELBO}

Standard VAE methodology is based on the bound
\begin{align*}
    \log p(\boldY) \ge \sum_{i=1}^N
    \mathbb{E}_{q_{\phi}(\boldz_i | \boldy_i)} [\log p_\theta(\boldy_i | \boldz_i)] - \text{KL}(q_{\phi}(\boldz_i|\boldy_i) || p(\boldz_i))
\end{align*}
However, unlike for standard VAE, for BasisVAE the $\log p_\theta(\boldy | \boldz)$ term is intractable due to integration over $\boldw$ and $\boldpi$. 
Now we apply the collapsing strategy of \citet{hensman_fast_2012, hensman_fast_2015} to $\log p_\theta(\boldy_i | \boldz_i)$ for all data points $i=1, \ldots, N$. 

Knowing that
\begin{align*}
     \log p_\theta(\boldy_i | \boldz_i, \boldpi)  &= \log \int  p_\theta(\boldy_i | \boldz_i, \boldpi, \boldw) p(\boldw) d\boldw \\
     &\ge \int q(\boldw) \log \frac{p_\theta(\boldy_i | \boldz_i, \boldw) p(\boldw | \boldpi)}{q(\boldw)} d\boldw \\
     &= \mathbb{E}_{q(\boldw)} \log \frac{p_\theta(\boldy_i | \boldz_i, \boldw) p(\boldw | \boldpi)}{q(\boldw)}
\end{align*}
we can now lower bound 
\begin{align*}
    \log p_\theta(\boldy_i | \boldz_i) &= \log \int p_\theta(\boldy_i | \boldz_i, \boldpi) p(\boldpi) d\pi \\
    &\ge \log \int \exp \left(
        \mathbb{E}_{q(\boldw)} \log \frac{p_\theta(\boldy_i | \boldz_i, \boldw) p(\boldw | \boldpi)}{q(\boldw)} 
    \right)  
    p(\boldpi) d\pi \\
    &= 
    \mathbb{E}_{q(\boldw)} \log p_\theta(\boldy_i | \boldz_i, \boldw) 
    + \log \int \exp \left( \mathbb{E}_{q(\boldw)} p(\boldw | \boldpi) \right) p(\boldpi) d\pi 
    - \mathbb{E}_{q(\boldw)} \log q(\boldw)
\end{align*}
where now all integrals can be calculated in closed form. Combining the two lower bounds, we obtain
\begin{align*}
    \log p(\boldY) \ge & 
    \sum_{i=1}^N
    \mathbb{E}_{q_{\phi}(\boldz_i | \boldy_i)} \mathbb{E}_{q(\boldw)} \log p_\theta(\boldy_i | \boldz_i, \boldw) - \sum_{i=1}^N \text{KL}(q_{\phi}(\boldz_i|\boldy_i) || p(\boldz_i)) \\
    &+ \log \int \exp \left( \mathbb{E}_{q(\boldw)} p(\boldw | \boldpi) \right) p(\boldpi) d\pi \\
    &- \mathbb{E}_{q(\boldw)} \log q(\boldw)
\end{align*}

Here, the first expected log-likelihood term can be calculated as
\begin{align*}
    \sum_{i=1}^N \mathbb{E}_{q_{\phi}(\boldz_i | \boldy_i)} \mathbb{E}_{q(\boldw)} \log p_\theta(\boldy_i | \boldz_i, \boldw) 
    &= \sum_{i=1}^N \mathbb{E}_{q_{\phi}(\boldz_i | \boldy_i)} \sum_{j=1}^P \sum_{k=1}^K \phi_{j,k} \log \mathcal{N}(y_i^{(j)} | \lambda_{j,k} f_{\text{basis}}^{(k)}(\boldz_i + \delta_{jk}), \; \sigma^2_j)
\end{align*}
The remaining challenging term is the third one, but it has a closed form as follows: knowing that
\begin{align*}
    \mathbb{E}_{q(\boldw)} p(\boldw | \boldpi)
    =\sum_{j=1}^P \sum_{k=1}^K \phi_{j,k} \log \pi_k
    = \sum_{k=1}^K n_k \log \pi_k
\end{align*}
where we have denoted $n_k := \sum_{j=1}^P \phi_{j,k}$, this term can now be expressed
\begin{align*}
    \log \int \exp \left( \mathbb{E}_{q(\boldw)} p(\boldw | \boldpi) \right) p(\boldpi) d\pi
    &= \log \int \exp \left( \sum_{k=1}^K n_k \log \pi_k \right) p(\boldpi) d\pi \\
    &= \log \int \prod_{k=1}^K \pi_k^{n_k}  \frac{1}{B(\boldalpha)} \pi_k^{\alpha_k - 1} d \boldpi \\
    &= \log \frac{1}{B(\boldalpha)} \int \prod_{k=1}^K  \pi_k^{n_k + \alpha_k - 1} d \boldpi \\
    &= \log B(\mathbf{n}+\boldalpha) - \log B(\boldalpha)
\end{align*}

where the normalising constant $B(\boldalpha) = \frac{\Gamma(\sum_k \alpha_k)}{\prod_k \Gamma(\alpha_k)}$ is the multivariate Beta function. 

Note that this hybrid inference scheme combines amortised reparameterisation-based inference for $\boldz$ and more classical approaches combined with collapsing for inference over $\boldw, \boldpi$.

\section{Adaptation of ELBO for large data sets} \label{supp:beta}

For large high-dimensional data sets (i.e.\ for large $N$ and $P$), the lower bound derived above will be dominated by the data log-likelihood. Also, the KL-term $\sum_{i=1}^N \text{KL}(q_{\phi}(\boldz_i|\boldy_i) || p(\boldz_i))$ can become large for increasing $N$. As a result, the clustering prior that we have introduced will implicitly become relatively less important when $N$ and $P$ increase. While the property that for large data sets the likelihood will dominate the prior is inherent to Bayesian models, it may not always be desirable, especially for mis-specified models. 

In practice, one way to alleviate this problem where the likelihood starts to dominate is via introducing weights that either downweigh the likelihood or upweigh the prior. For example, $\beta$-VAE modifies the standard VAE lower bound by scaling the KL term by a constant $\beta>0$
\citep{higgins_beta-vae:_2017}. 
Even though the resulting expression is not a lower bound on the original log marginal likelihood any more, it is closely connected to an ELBO on an alternative formulation with an annealed prior $p(\boldz)^\beta / \int p(\boldz)^\beta d \boldz$ \citep{hoffman_beta-vaes_2017, mathieu_disentangling_2019}.  
Analogously to the $\beta$-VAE approach, we propose to achieve a similar effect, by introducing $\beta, \gamma$ as part of the following objective
\begin{align*}
    \calL_{\beta, \gamma} := 
    \sum_{i=1}^N &
    \mathbb{E}_{q_{\phi}(\boldz_i | \boldy_i)} \mathbb{E}_{q(\boldw)} \log p_\theta(\boldy_i | \boldz_i, \boldw) - \beta \sum_{i=1}^N \text{KL}(q_{\phi}(\boldz_i|\boldy_i) || p(\boldz_i)) \\
    &+ \gamma \left( \log \int \exp \left( \mathbb{E}_{q(\boldw)} p(\boldw | \boldpi) \right) p(\boldpi) d\pi
    - \mathbb{E}_{q(\boldw)} \log q(\boldw) 
     \right)
\end{align*}
Note that the setting $\beta=1, \gamma=1$ corresponds to the original ELBO, whereas 
\begin{itemize}
    \item $\beta > 1$ provides us with a handle to increasing the relative importance of the prior $p(\boldz_i)$, to compensate for large data dimensionality $P$
    \item $\gamma > 1$ provides us with a mechanism increase the relative importance of the sparse clustering prior, compensating for a combination of large sample size $N$ and data dimensionality $P$
\end{itemize}
For the large-scale applications we consider here, we propose to choose $\beta=\gamma$, i.e.\ the following bound $\calL_\beta$
\begin{align*}
    \calL_{\beta} := 
    \sum_{i=1}^N &
    \mathbb{E}_{q_{\phi}(\boldz_i | \boldy_i)} \mathbb{E}_{q(\boldw)} \log p_\theta(\boldy_i | \boldz_i, \boldw) - \beta \sum_{i=1}^N \text{KL}(q_{\phi}(\boldz_i|\boldy_i) || p(\boldz_i)) \\
    &+ \beta \left( \log \int \exp \left( \mathbb{E}_{q(\boldw)} p(\boldw | \boldpi) \right) p(\boldpi) d\pi
    - \mathbb{E}_{q(\boldw)} \log q(\boldw) 
     \right)
\end{align*}
In our experiments, for moderately sized synthetic data (where $P=50$) we used $\beta=1$, whereas for single-cell gene expression data (where $P=5000$) we used $\beta=20$. 

\section{Single-cell likelihood}

For non-negative valued single-cell data, we replace the Gaussian likelihood in the decoder with a zero-inflated negative binomial (ZINB) likelihood. This is a common approach due to the data being non-negative-valued and exhibiting zero-inflation, see e.g.\ \citep{lopez_deep_2018}. 

We first introduce notation for the negative binomial distribution as follows
\begin{align*}
    y_i^{(j)} \sim \text{NB}(\mu_{ij}, \phi_{ij}) \iff 
    p(y_i^{(j)}) 
    = \frac{\Gamma(y_i^{(j)}+\phi_{ij})}{\Gamma(y_i^{(j)}+1)\Gamma(\phi_{ij})} 
    \left( \frac{\phi_{ij}}{\phi_{ij}+\mu_{ij}} \right)^{\phi_{ij}}
    \left( \frac{\mu_{ij}}{\phi_{ij}+\mu_{ij}} \right)^{y_i^{(j)}}
\end{align*}
and for ZINB
\begin{align*}
    y_i^{(j)} \sim \text{ZINB}(\mu_{ij}, \phi_{ij}, \pi_{ij}) \iff 
    p(y_i^{(j)})  = \pi_{i, j} \delta_0 + (1-\pi_{i, j}) \text{NB}( \mu_{ij} , \phi_{ij}) .
\end{align*}
which is a mixture of a negative binomial distribution and a point mass at zero. Here $\pi_{i, j}$ are referred to as the dropout probabilities. 
Now we outline the ZINB likelihood for a data vector $y_i^{(1:P)} \in \R^P$ in our generative model
\begin{align*}
    y_i^{(1:P)} | \boldz_i, \phi \sim \pi^{1:P}_{\psi}(\boldz_i) \cdot \delta_0 + (1-\pi^{1:P}_{\psi}(\boldz_i)) \cdot \text{NB}( \text{softplus}(f^{1:P}_\theta(\boldz_i)) , \phi^{1:P}) .
\end{align*}
where now both $\pi^{1:P}_{\psi}$ and $f^{1:P}_{\theta}$ are decoding neural networks (i.e.\ we parameterise both the mean and the dropout probabilities via neural networks) and $\phi^{1:P}$ are the inverse dispersion parameters of the negative binomial distribution. 
As we are mainly interested in clustering features based on the \emph{mean} function, we decided to apply BasisVAE clustering structure only within the decoder network $f^{1:P}_{\theta}$ (thus $\pi^{1:P}_{\psi}$ is just a standard neural network). 

\section{Additional Figures}

\subsection{Additional figures for synthetic data} \label{supp:fig_synthetic}

\begin{figure}[!ht]
    \centering
    \includegraphics[width=\columnwidth]{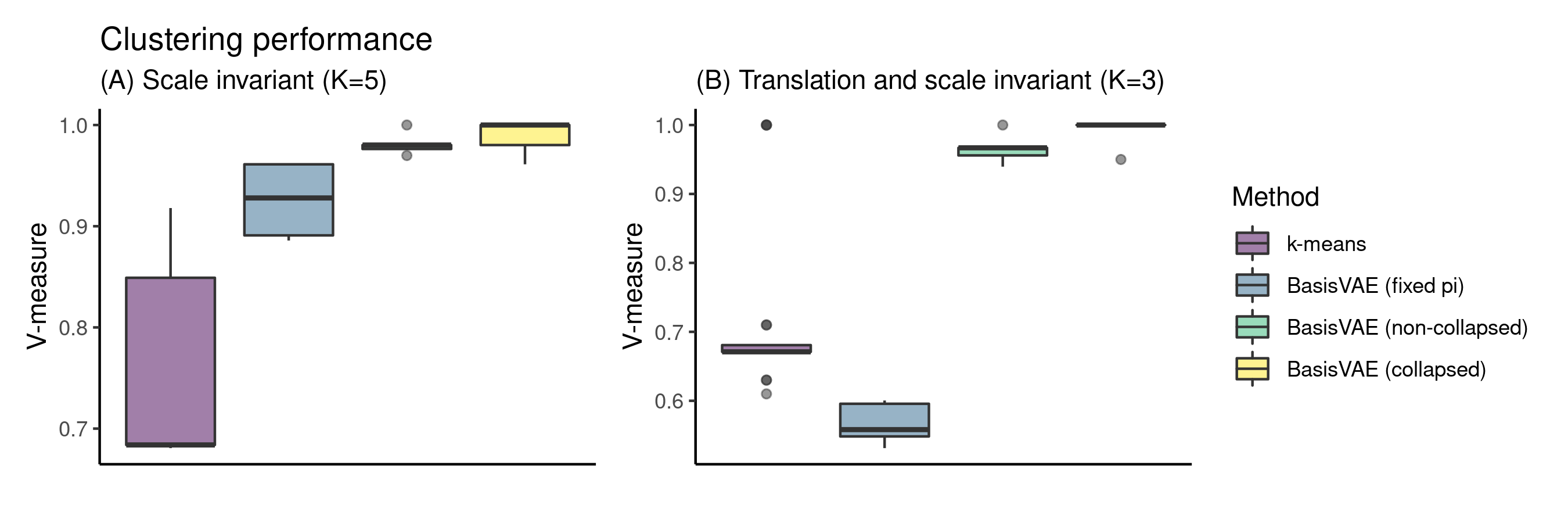}
    \caption{
        Using the V-measure to quantify the clustering performance of k-means and different versions of BasisVAE (with different inference techniques for $q(\boldw, \boldpi)$) on the synthetic data from Figure~\ref{fig:illustration}. Results are shown for both (A) scale invariant and (B) translation and scale invariant setting. For k-means, we used the true number of clusters, i.e.\ (A) $K=5$ and (B) $K=3$. For BasisVAE, we used an overspecified $K=20$. 
    }
    \label{fig:V_measure_toy}
\end{figure}

\subsection{Additional figures for single-cell Spermatogenesis data} \label{supp:fig_singlecell}

\begin{figure}[!th]
    \centering
    \includegraphics[width=\textwidth]{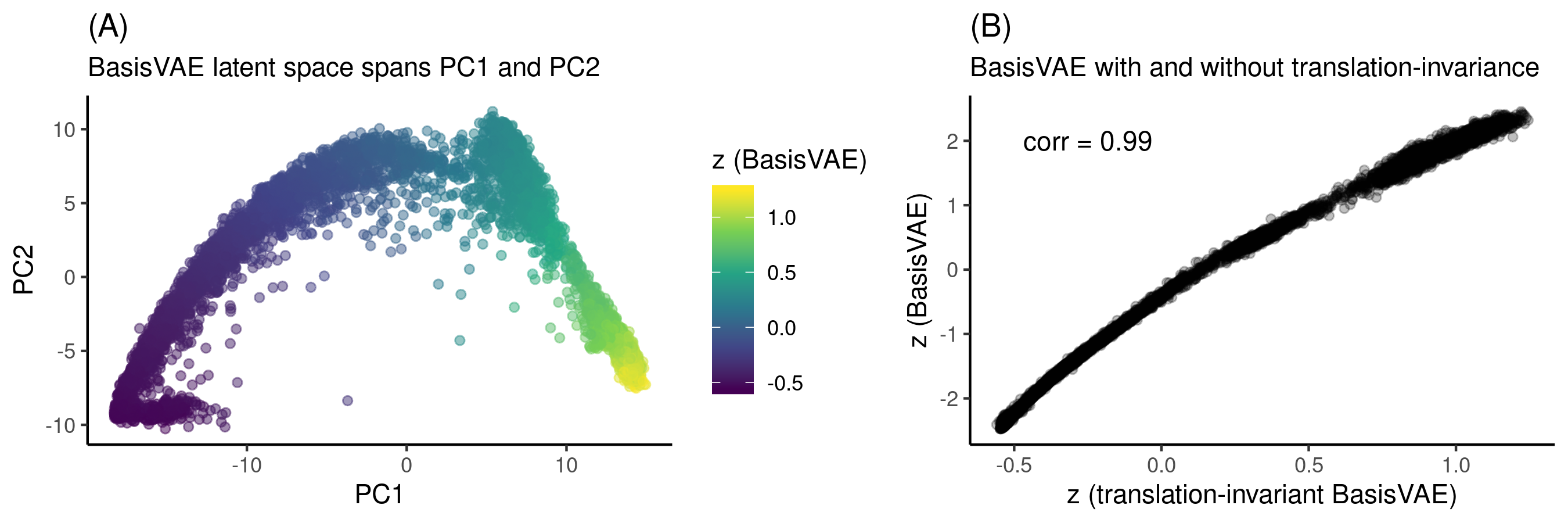}
    \caption{
        (A) BasisVAE has inferred a latent $\boldz \in \R$ that captures a trajectory in the (PC1, PC2) space. 
        (B) BasisVAE and its translation-invariant version infer a highly similar latent space (the respective inferred latent coordinates are highly correlated). 
    }
    \label{fig:spermatogenesis_latent_space}
\end{figure}

\end{appendices}

\end{document}